\documentclass[conference]{IEEEtran}
\usepackage[pdftex]{graphicx}
\usepackage{graphicx}
\usepackage{amsmath}
\usepackage{fixltx2e}
\usepackage{multirow}
\usepackage{textcomp}

\ifCLASSINFOpdf
\else
\fi
\usepackage{array}

\usepackage{mdwmath}
\usepackage{mdwtab}

\usepackage{eqparbox}

\usepackage[tight,footnotesize]{subfigure}
\begin{document}

\title{A Study of Morphological Filtering Using Graph and Hypergraphs}

\author{\IEEEauthorblockN{Keerthana S Prakash}
\IEEEauthorblockA{PG Student\\
 Computer Science and Engineering\\
College Of Engineering\\Karunagappally, India\\keerthanatheertha@gmail.com}
\and
\IEEEauthorblockN{Prakash R P}
\IEEEauthorblockA{PG Student\\ Computer Science and Engineering\\
	College Of Engineering\\Karunagappally, India\\prakasharumanoor@gmail.com}
\and
\IEEEauthorblockN{Binu V P}
\IEEEauthorblockA{Associate Professor
	\\Computer Science and Engineering\\
	College Of Engineering\\Karunagappally, India\\
binuvp@gmail.com}
}


%
%
%


\maketitle

\begin{abstract}
\boldmath
Mathematical morphology (MM) helps to describe and analyze shapes using set theory. MM can be effectively applied to binary images which are treated as sets. Basic morphological operators defined can be used as an effective tool in image processing. Morphological operators are also developed based on graph and hypergraph. These operators have found better performance
and applications in image processing. Bino et al. \cite{8}, \cite{9} developed the theory of morphological operators on hypergraph. A hypergraph structure is considered and basic morphological operation erosion/dilation is defined. Several new operators opening/closing and filtering are also defined on the hypergraphs. Hypergraph based filtering have found comparatively better performance with morphological filters based on graph. In this paper we evaluate the effectiveness of hypergraph based ASF on binary images. Experimental results shows that hypergraph based ASF filters have outperformed graph based ASF.

\end{abstract}
%
		
	




%
\IEEEpeerreviewmaketitle

\section{Introduction}
Mathematical morphology(MM) \cite{1}, \cite{2}, \cite{10}, \cite{12} is set theoretic based approach which is developed in 1960. It was extended to gray-level images in the late 1970's. This technique is used to extract characteristic features of the images which are useful for specific applications. Morphological operations are developed on graphs \cite{4}, \cite{5}, \cite{6} and hypergraph \cite{7}, \cite{8}, \cite{9}. Vincent\cite{5} proposed morphological operators on graph. Cousty et al. [6] defined morphological filtering operators on lattice of graph. Hypergraph based morphology is found more prominent compared with graph based morphology and can be used in several applications. Bino et al \cite{9}, \cite{9} proposed a hypergraph based morphological filter. Here we evaluate the effectiveness of hypergraph based ASF on binary images.\par
Graph consists vertices and edges. The collection of points are called vertices. The binary relation between them is usually represented by a subset $E\subseteq V \times V$ called the edges; The vertices v and w are related if and only if $(v,w)\in E.$  \par
Hypergraph theory, which is developed by C. Berge \cite{3} in 1960. Hypergraph \cite{7} is a generalized version of graphs where edges can connect any number of vertices and are called hyperedges. Hypergraph is defined as $H=(H^{\bullet},H^{\times})$. The vertices are set of points and represented as $H^{\bullet}$. The family of subset of $H^{\bullet}$ called hyperedges and represented as $H^{\times}$. The number of vertices  that are connected by an hyperedge determines the cardinality. Uniform hypergraphs where all hyperedges have the same cardinality are found useful applications image processing. Figure 1 gives the example of sample graph and hypergraph.

Alternating Sequential Filters \cite{6}, \cite{8}, \cite{9}, \cite{11} are composition of openings and closings which form granulometric families of increasing sizes. ASFs are also developed on these structures. We evaluate the performance of hypergraph based filters using this technique. 
\par
Contents of this paper as given. In Section 2, we introduces the morphology operators on graph, recall basic morphological operators, new openings/closings on hypergraph, explains granulometries and alternate sequential filters works on hypergraph and illustrates the idea for generating hypergraph from images. Section 2, we evaluate the performance of hypergraph based ASF filters on binary image. In Section 3, experimental results are given. Section 4 contains conclusion of the paper and the future works in this field.
\section{Preliminaries}
\subsection{Morphological Operators on Graphs}
Graphs can be used to represent the structural information about the elements in the digital objects. Cousty et al. \cite{6} proposed basic operators which helps to develop dilation/erosion, half openings/half closings, granulometries and alternate sequential filters acting on graphs.
\begin{figure}[!htb]\centering
	\begin{minipage}{0.20\textwidth}
		\includegraphics[width=\linewidth]{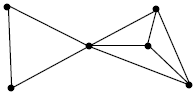}
		\label{fig1}
	\end{minipage}
	\hspace{0.6cm}
	\begin {minipage}{0.20\textwidth}
	\includegraphics[width=\linewidth]{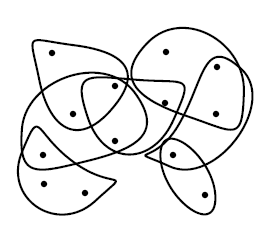}
	\label{fig2}
\end{minipage}
\caption{Example of sample graph and hypergraph}
\end{figure} 
\subsection{Morphological Operators on Hypergraph}
Complex relations in the images can be represented by using hypergraph. Bino et al. \cite{8}, \cite{9} defined morphological operators on hypergraph. Hypergraph consist of sets of points and sets of hyperedges. The basic operators are considered which is derived from vertex set to the edge set and vice versa. \\
\\
\textbf{Property 1}. \cite{8}, \cite{9} For any $ X^{\bullet} \subseteq H^{\bullet} $ and any $ X^{\times} \subseteq H^{\times} $, where $ X^{\times}=(e_{j} ) $, $j \in J$
such that $ J\subseteq I $\\
\begin{enumerate}
	\item  $\delta^{\bullet} \colon H^{\times} \rightarrow H^{\bullet}$ is such that $ \delta^{\bullet}(X^{\times})=\underset{j \in J}{\cup} v(e_{j}).$
	\item{ $\epsilon^{\times}\colon H^{\bullet} \rightarrow H^{\times}$  is such that \\$ \epsilon^{\times}(X^{\bullet}) =\{{e_{i},i \in I|v(e_{i})\subseteq X^{\bullet}}.$}\}
	\item {$ \epsilon^{\bullet}\colon H^{\times} \rightarrow H^{\bullet} $ is such that $ \epsilon^{\bullet}(X^{\times})=\underset{j\notin J}{\cap} 
		\overline{v(e_{j})}$}. 
	\item{ $\delta^{\times}\colon H^{\bullet} \rightarrow H^{\times}$  is such that\\ $ \delta^{\times}(X^{\bullet}) =\{{e_{i},i \in I|v(e_{i}) \cap X^{\bullet}\neq\emptyset}$}\}.
	\\
	
\end{enumerate}
\textbf{Property 2. \cite{6}, \cite{8}, \cite{9} (dilation, erosion, adjunction, duality)}\\
\begin{enumerate}
	\item {Operators $\epsilon^{\times}$ and $\delta^{\times}$ (resp. $\epsilon^{\bullet}$ and $\delta^{\bullet})$ are dual of each other}.
	\item Both $(\epsilon^{\times}$, $\delta^{\bullet})$ and $(\epsilon^{\bullet},\delta^{\times})$ are adjunctions.
	\item Operators $\epsilon^{\bullet}$ and $\epsilon^{\times}$ are erosions.
	\item Operators $\delta^{\bullet}$ and $\delta^{\times}$ are dilations.
	\\
\end{enumerate}
\textbf{Definition 1. \cite{6}, \cite{8}, \cite{9} (vertex dilation, vertex erosion)}
$\delta$ and $\epsilon$ defined that act on $H^{\bullet}$ by $\delta=\delta^{\bullet}\circ\delta^{\times}$ and $\epsilon=\epsilon^{\bullet}\circ\epsilon^{\times}$.\\
\\
\textbf{Property 3. \cite{8}, \cite{9}, \cite{14} For any $X^{\bullet} \subseteq H^{\bullet}$}\\
\begin{enumerate}
	\item $\delta(X^{\bullet})=\{x\in H^{\bullet}| \exists e_{i}, i\in I$ such that \\$x\in v(e_{i})$ and $v(e_{i})\bigcap X^{\bullet}\neq\emptyset\}.$
	\item $\epsilon(X^{\bullet})=\{x\in H^{\bullet}| \exists e_{i}, i\in I$ such that \\$x\in v(e_{i})$, $v(e_{i})\subseteq X^{\bullet}\}.$
	\\
\end{enumerate}
\textbf{Definition 2. \cite{8}, \cite{9} (hyper-edge dilation, hyper-edge erosion)} $\Delta$ and $\varepsilon$ defined that act on $H^{\times}$ by $\Delta=\delta^{\times}\circ\delta^{\bullet}$ and $\varepsilon=\epsilon^{\times}\circ\epsilon^{\bullet.}$
\\
\\
\textbf{Property 4.} \cite{8}, \cite{9} For any $X^{\times}\subseteq H^{\times}, X^{\times}=(e_{j})_{j\in J}$
\begin{enumerate}
	\item $\Delta(X^{\times})$=$\{e_{i},i\in I|\exists e_{j},j\in J$ such that \\
	$ v(e_{i})\cap v(e_{j}\neq\emptyset)\}.$
	
	\item $\varepsilon(X^{\times})=\{e_{j},j\in J|v(e_{j})\cap v(e_{i})\neq\emptyset,\forall i\in I, J \}.$
	\\
\end{enumerate}
\textbf{Definition 3. \cite{8}, \cite{9} (hypergraph dilation, hypergraph erosion)} The operators $[\delta,\Delta]$ and $[\epsilon,\varepsilon]$ defined by respectively $[\delta,\Delta](X)=(\delta(X^{\bullet}),\Delta(X^{\times}))$ and $[\epsilon,\varepsilon](X)=(\epsilon(X^{\bullet}),\varepsilon(X^{\times}) ),$ for any $X\in H.$\\\\\textbf{Definition 4. \cite{9}, \cite{13} (opening, closing)}
\begin{enumerate}
	\item $\gamma_{1}$ and $\varPhi_{1}$ defined, that act on $H^{\bullet}$, by $\gamma_{1}=\delta \circ \epsilon$ and $\varPhi_{1}=\epsilon \circ \delta.$
		
	\item $\Gamma_{1}$ and $\Phi_{1}$ defined, that act on $H^{\times}$, by $\Gamma_{1}=\Delta\circ\varepsilon$ and $\Phi_{1}=\varepsilon \circ
		\Delta.$
	\item $[\gamma,\Gamma]_{1}$ and $[\varPhi,\Phi]_{1}$ defined, that act on H by respectively $[\gamma,\Gamma]_{1}(X)=(\gamma_{1}(X^{\bullet}),\Gamma_{1}(X^{\times}))
		$and $[\varPhi,\Phi]_{1}(X)=(\varPhi_{1}(X^{\bullet}),\Phi_{1}(X^{\times}))$ for any $X\in H.$
		
\end{enumerate}	

\textbf{Definition 5. \cite{9}, \cite{13} (half-opening, half-closing)}\\
\begin{enumerate}
	\item $\gamma_{1/2}$ 
		and $\varPhi_{1/2}$ defined, that act on $H^{\bullet}$, by $\gamma_{1/2}=\delta^{\bullet} \circ \epsilon^{\times}$ and $\varPhi_{1/2}=\epsilon^{\bullet} \circ \delta^{\times}.$
		
	\item $\Gamma_{1/2}$ and $\Phi_{1/2}$ defined, that act on $H^{\times}$, by 
		$\Gamma_{1/2}=\delta^{\times}\circ\epsilon^{\bullet}$ and $\Phi_{1/2}=\epsilon^{\times} \circ
		\delta^{\bullet}.$
		\\
\end{enumerate}
\textbf{Property 5. \cite{9}, \cite{13} (hypergraph opening, hypergraph closing)}\\
	\begin{enumerate}
		\item The operators $\gamma_{1/2}$ and $\gamma_{1}$ (resp. $\Gamma_{1/2}$ and $\Gamma_{1})$ are opening on $H^{\bullet}$ (resp. $H^{\times}$) and $\varPhi_{1/2}$ and $\varPhi_{1}$ (resp. $\Phi_{1/2}$ and $\Phi_{1})$ are closing on $H^{\bullet}$.
		
		\item The family H is closed under $[\gamma,\Gamma]_{1/2},[\varPhi,\Phi]_{1/2},$ $[\gamma,\Gamma]_{1}$ and $[\varPhi,\Phi]_{1}.$
		\item $[\gamma,\Gamma]_{1/2}$ and $[\gamma,\Gamma]_{1}$ are opening on H and $[\varPhi,\Phi]_{1/2}$ and $[\varPhi,\Phi]_{1}$ are closing on H.
		
	\end{enumerate}
\subsection{Alternative Sequential Filters}

Granulometries \cite{6}, \cite{9} contains families of openings and closings that are parametrized by a positive number.\\
\\
\textbf{Definition 6.} \cite{9} Let $\lambda \in N$. $[\gamma,\Gamma]_{\lambda/2}$ (resp. $[\varPhi,\Phi]_{\lambda/2}$) defined as follows. $[\gamma,\Gamma]_{\lambda/2}=[\delta,\Delta]^{i} \circ (\gamma,\Gamma]_{1/2})^{j} \circ [\epsilon,\varepsilon]^{i},$ where i and j are respectively the quotient and reminder when $\lambda$ is divided by 2.
\\
\\
\textbf{Definition 7.} \cite{9}
Let $\lambda \in N$ and $X\in H$. $ASF_{\lambda/2}$ is equal to X if $\lambda=0$ and $[\gamma,\Gamma]_{\lambda/2}\circ[\varPhi,\Phi]_{\lambda /2}\circ ASF_{(\lambda-1)/2}(X) $ if $\lambda \neq 0$
\\
\begin{equation*}
ASF_{\lambda/2}(X)=
\begin{cases}
X \;\;, if \;\lambda=0\\
[\gamma,\Gamma]_{\lambda/2}\circ[\varPhi,\Phi]_{\lambda/2}\circ ASF_{(\lambda-1)/2} (X)\;\;, if \lambda \neq 0

\end{cases}
\\\\
\\
\end{equation*}

\subsection{Generating Hypergraph from Images}

To consider morphological operators on binary image, a hypergraph is created based on the image. Initially vertices are created corresponding to each pixel. Then hyperedges are placed among collection of vertices. A variety of methods can be used to form hyperedges. Figure 2 give a method to construct hypergraph where each vertex belong to exactly four hyperedges \cite{8}, \cite{9}.
\begin{figure}[h]
	\centering
	\includegraphics[width=0.22\textwidth]{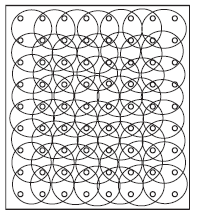}
	\caption{Hyperedges forming four uniform hypergraphs}
	\label{fig:}
\end{figure}
\section{Experimental Results}

To analyse the working of filters on the binary image, we consider the binary image of lena with size $1024\times1024$ which is resized from the image of size $512\times512$ in Figure 3. We use mean square error as the error measure. The noisy image shows in Figure 4 which has MSE equal to 10$\% $.
\\
\\   
\begin{figure}[!htb]\centering
	\begin{minipage}{0.20\textwidth}
		\includegraphics[width=\linewidth]{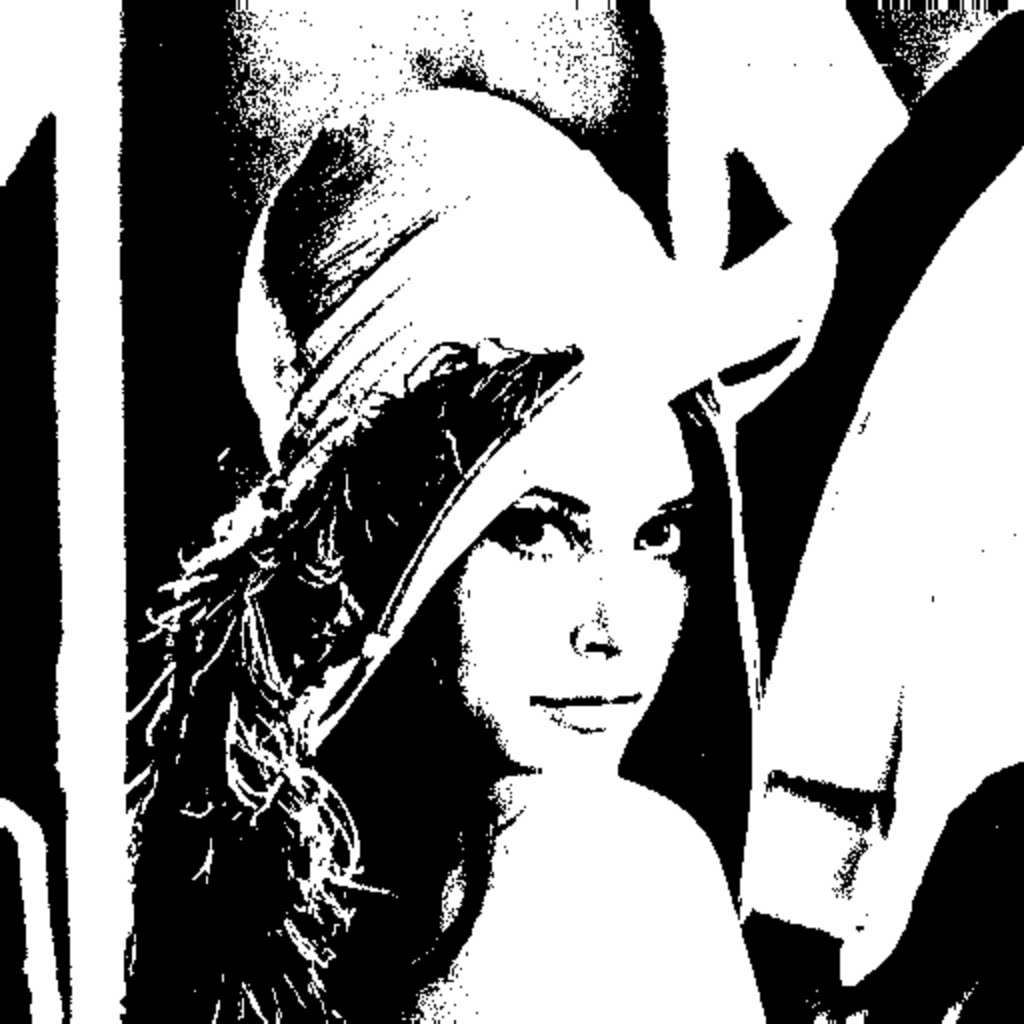}
		\caption{Original Image}
		\label{fig1}
	\end{minipage}
	\hspace{0.6cm}
	\begin {minipage}{0.20\textwidth}
	\includegraphics[width=\linewidth]{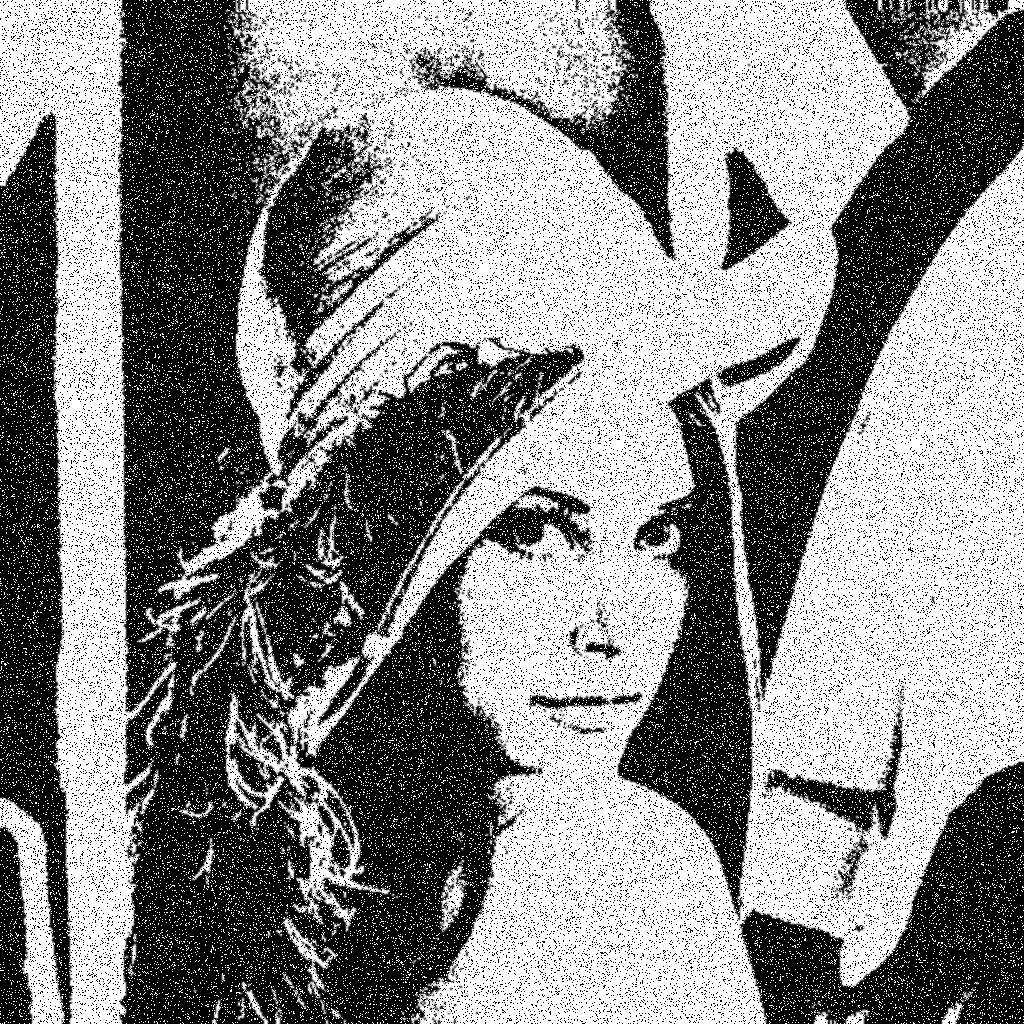}
	
	\caption{Noicy Image, MSE = 10$\% $}

\end{minipage}
\end{figure}

First we form a 4 uniform hypergraph based on the image. Then noisy image shown in Figure 4 was then processed, using the alternative sequential filters upto filters of size 7. The graph based ASF gives minimum mean square error of 1.99$\%$ for the best result, shown in Figure 5. But hypergraph based ASF gives minimum mean square error of 1.76$\%$ for the best result, which is shown in Figure 6. Hypergraph based ASF gives minimum error value compared graph ASF.\\
\\

\par

When increasing the scale of ASF, the mean square error is first decreasing, reaches a minimum value and then increasing. There is an increase in mean square error after reaching the minimum value since holes are filled instead of noise removal. This can be visualized by plotting error measure versus different values of $\lambda$ for graph ASF and hypergraph ASF. Figure 7 and Figure 8 shows the respective plots.

\begin{figure}[!htb]\centering
	\begin{minipage}{0.20\textwidth}
		\includegraphics[width=\linewidth]{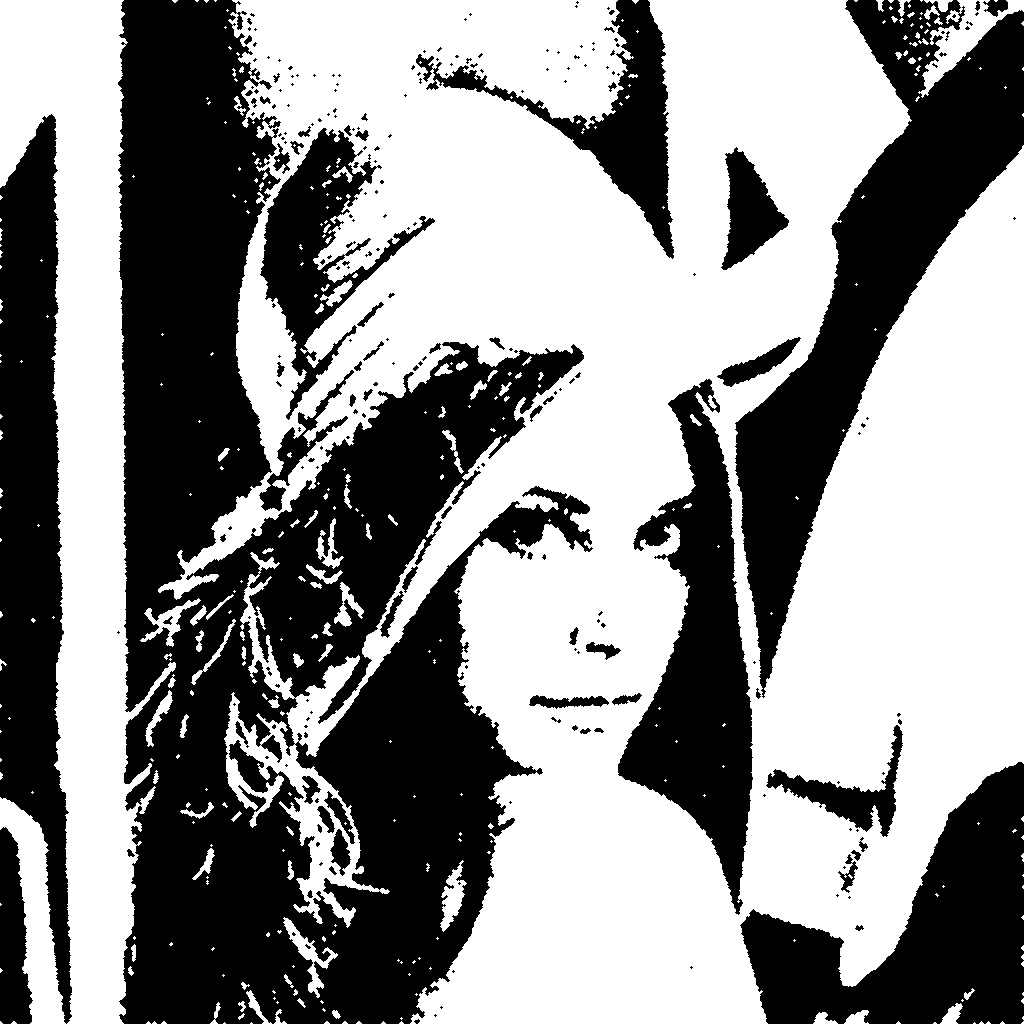}
		\caption{Result of graph ASF}
		\label{fig1}
	\end{minipage}
	\hspace{0.6cm}
	\begin {minipage}{0.20\textwidth}
	\includegraphics[width=\linewidth]{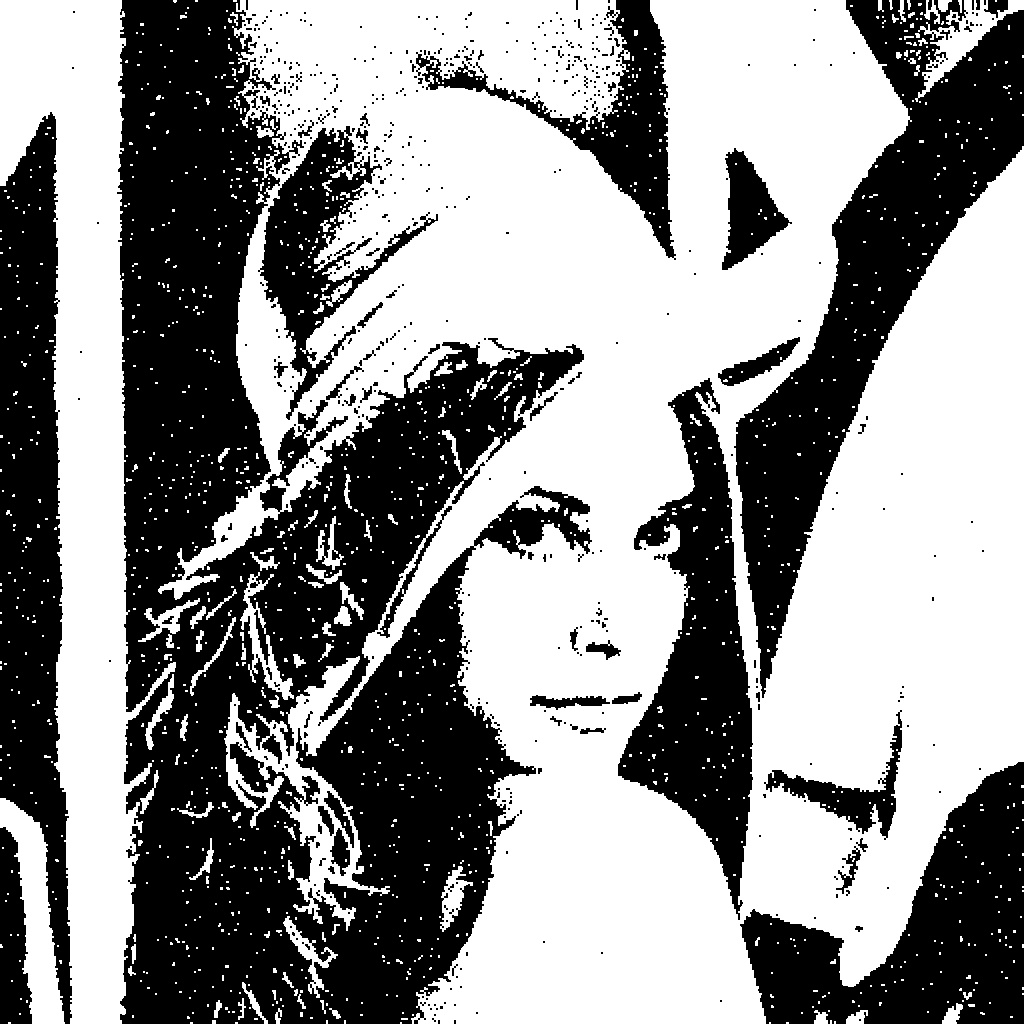}
	\caption{Result of Hypergraph ASF}
	\label{fig2}
\end{minipage}
\end{figure}

\begin{figure}[!htb]\centering
	\begin{minipage}{0.20\textwidth}
		\includegraphics[width=\linewidth]{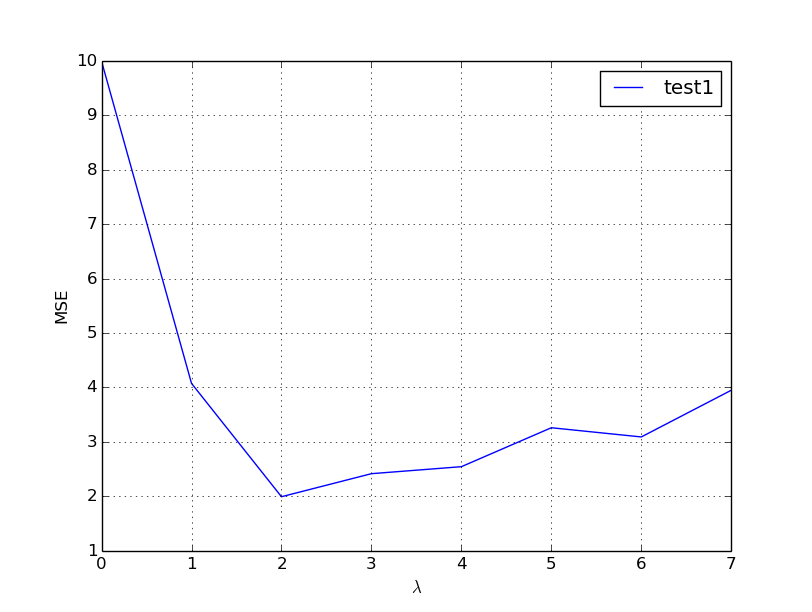}
		\caption{Mean Square Error value versus size of the filter in graph ASF}
		\label{fig1}
	\end{minipage}
	\hspace{0.6cm}
	\begin {minipage}{0.20\textwidth}
	\includegraphics[width=\linewidth]{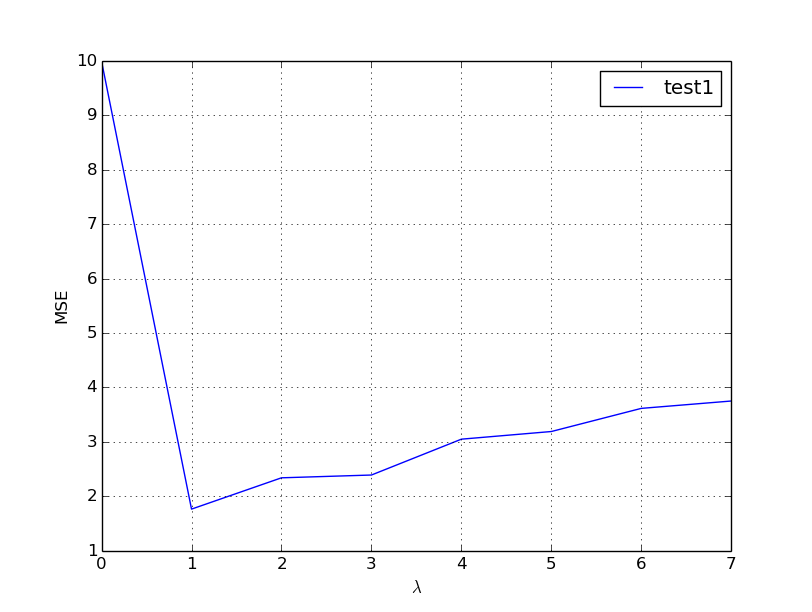}
	\caption{Mean Square Error value versus size of the filter in hypergraph ASF}
	\label{fig2}
\end{minipage}
\\
\end{figure}

To evaluate the filtering results quantitatively, we take a binary image of lena $(1024\times1024)$. Then take four sets of noisy versions of that particular image by adding different level of salt and pepper noise which is given in figure 9. Each images are filtered using  median filter, graph ASF and hypergraph ASF. The best results of each filter for all noisy version are taken. The neighborhood size of the median filter and $\lambda$ value of ASF filters which shows minimum mean square error is also specified. The mean square error calculated in percentage and shown in Table 1. 
\\
\par The same operations are applied for a  MRI image of size $1125\times1125$ which is resized from the image of size $225\times225$. The noisy versions of the image shows in Figure 10. The results are given in Table 2. The tables shows that hypergraph ASF perform well than other filters. But, when increasing the level of noise, the error in hypergraph based ASF increases. Morphological filters performs well on high resolution images. So median filter give minimum error in images with higher level of noise. But Better results can be obtained by applying ASF filter with different set of hyperedges.  \par
\begin{figure}[!htb]\centering
	\hspace{0.6cm}
	\begin{minipage}{0.10\textwidth}
		\includegraphics[width=\linewidth]{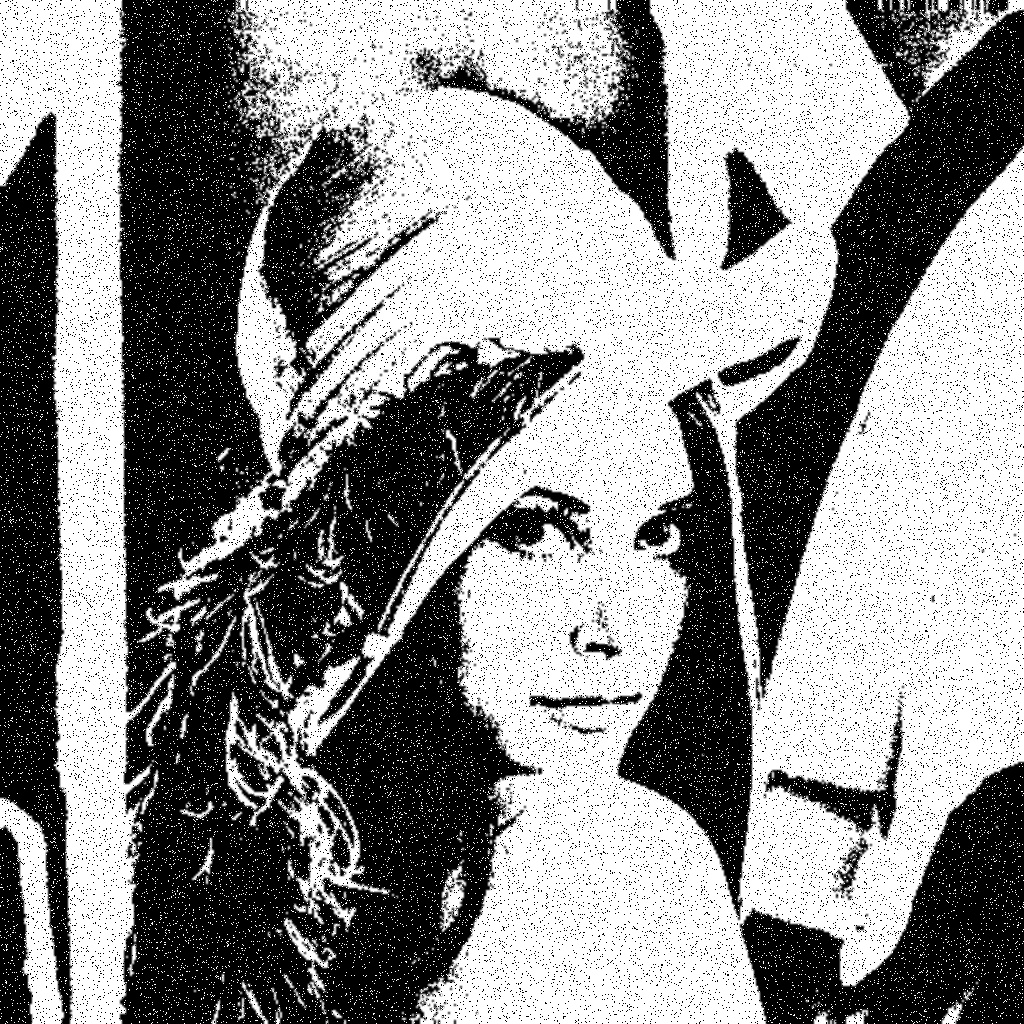}
		\label{fig1}
	\end{minipage}
	\hspace{0.6cm}
	\begin {minipage}{0.10\textwidth}
	\includegraphics[width=\linewidth]{10}
	\label{fig2}
\end{minipage}

\hspace{0.6cm}
\begin {minipage}{0.10\textwidth}
\includegraphics[width=\linewidth]{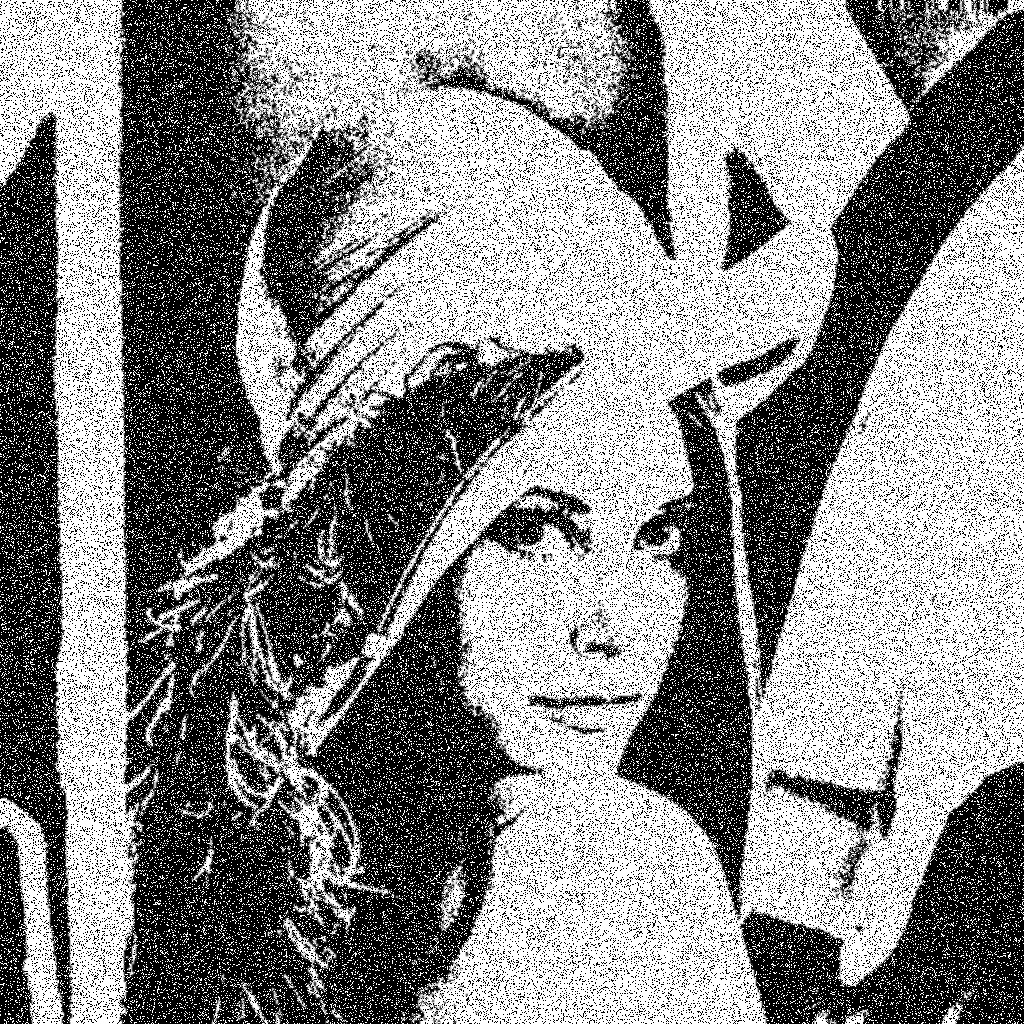}
\label{fig2}
\end{minipage}
\hspace{0.6cm}
\begin {minipage}{0.10\textwidth}
\includegraphics[width=\linewidth]{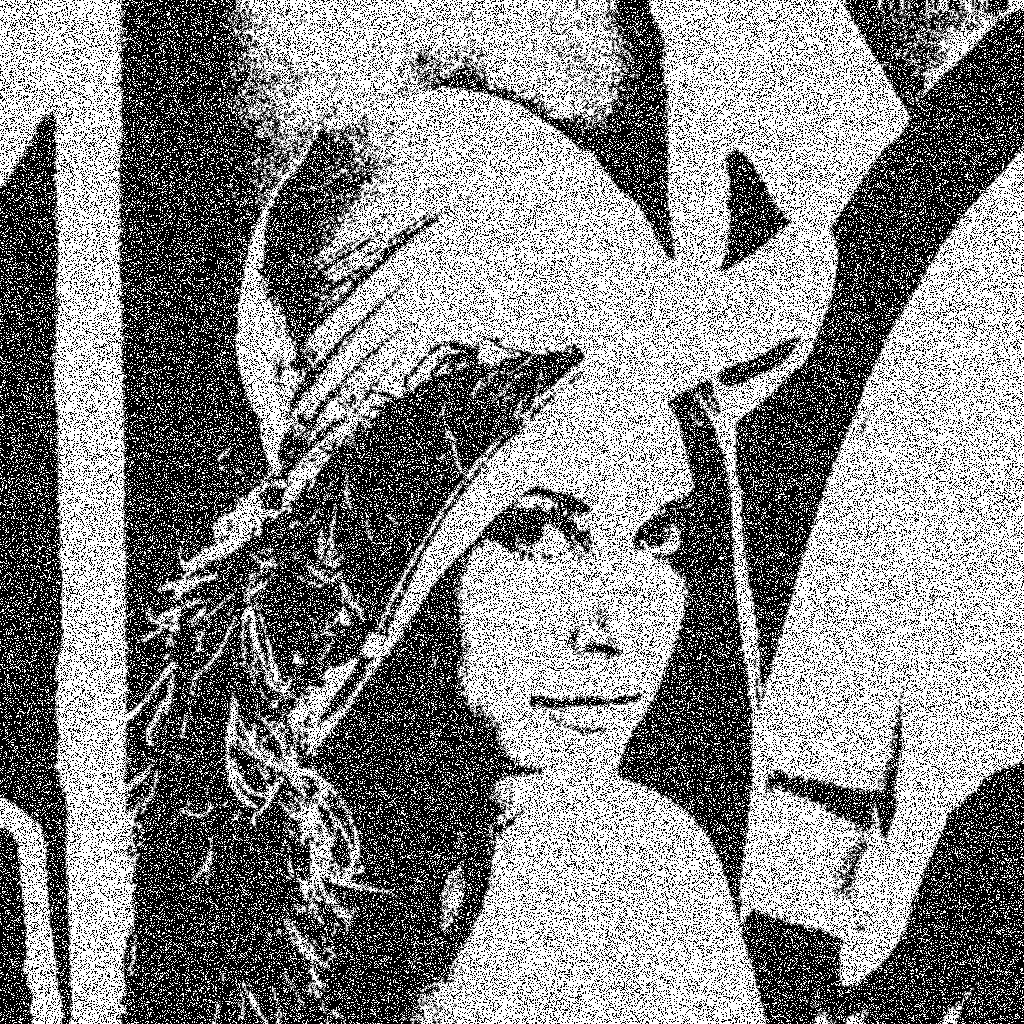}
\label{fig2}
\end{minipage}
\caption{Images of lena with increasing level of noise}
\end{figure}

\begin{center}
	TABLE 1\\
	MSE FOR DIFFERENT TYPES OF FILTERS APPLIED TO FOUR NOISY VERSIONS OF LENA IMAGES 
\end{center}
\begin{center}
	\begin{tabular}{|c|c|c|c|}

		\hline
		
		Noise Ratio&Median Filter&Graph
		ASF&Hypergraph ASF\\
		
		\hline
		
		5 & 1.37 $(3\times3)$ & 1.44(1) & 0.51(1)\\
		10 & 1.58 $(3\times3)$ & 1.99(2) & 1.76(1)\\
		15 & 2.27 $(3\times3)$ & 2.71(2) & 3.19(3)\\
		20 & 2.82 $(5\times5)$ & 3.55(4) & 5.15(5)\\
		\hline
		
	\end{tabular}
	\\
	
\end{center}

\begin{figure}[!htb]\centering
	\hspace{0.6cm}
	\begin{minipage}{0.10\textwidth}
		\includegraphics[width=\linewidth]{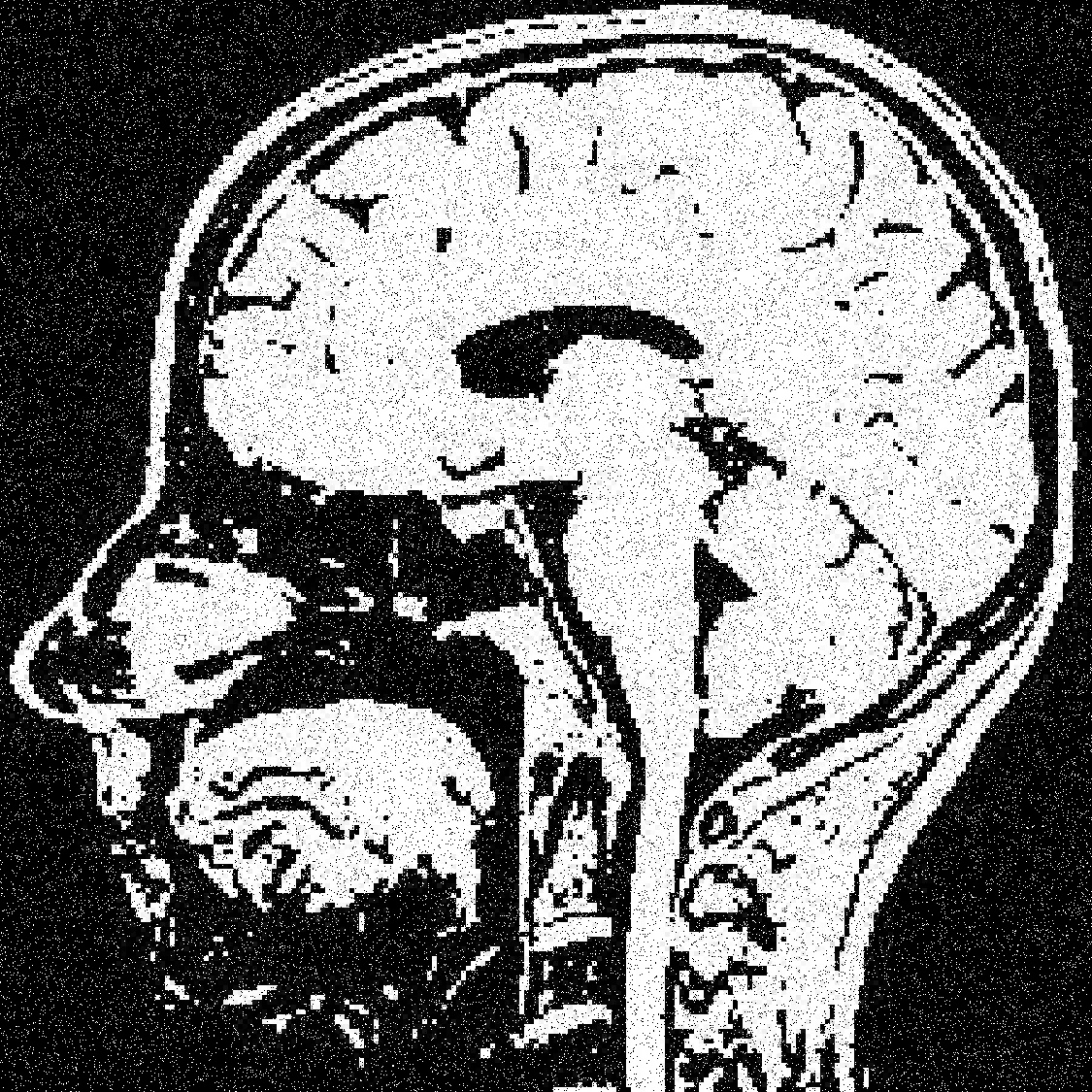}
		\label{fig1}
	\end{minipage}
	\hspace{0.6cm}
	\begin {minipage}{0.10\textwidth}
	\includegraphics[width=\linewidth]{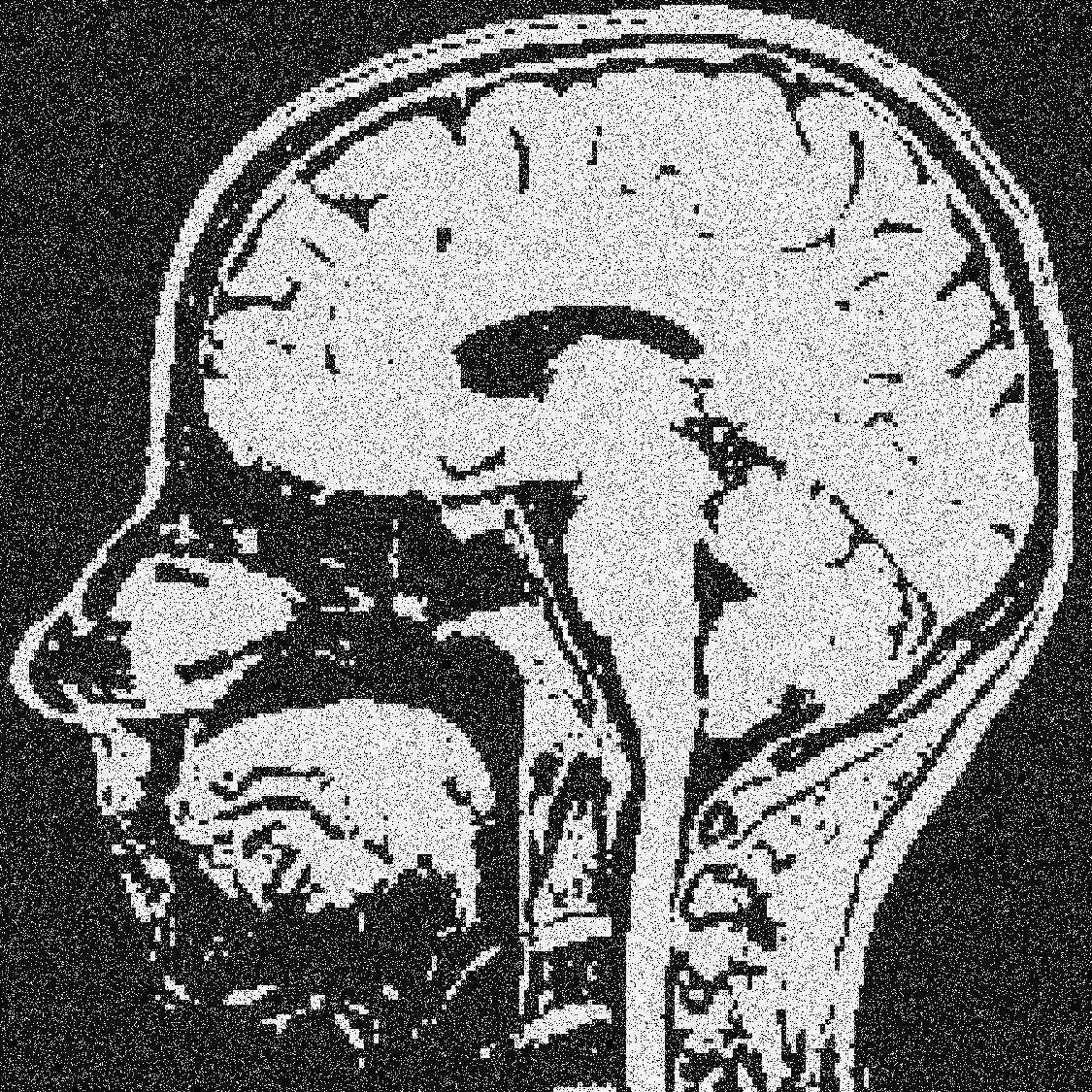}
	\label{fig2}
\end{minipage}

\hspace{0.6cm}
\begin {minipage}{0.10\textwidth}
\includegraphics[width=\linewidth]{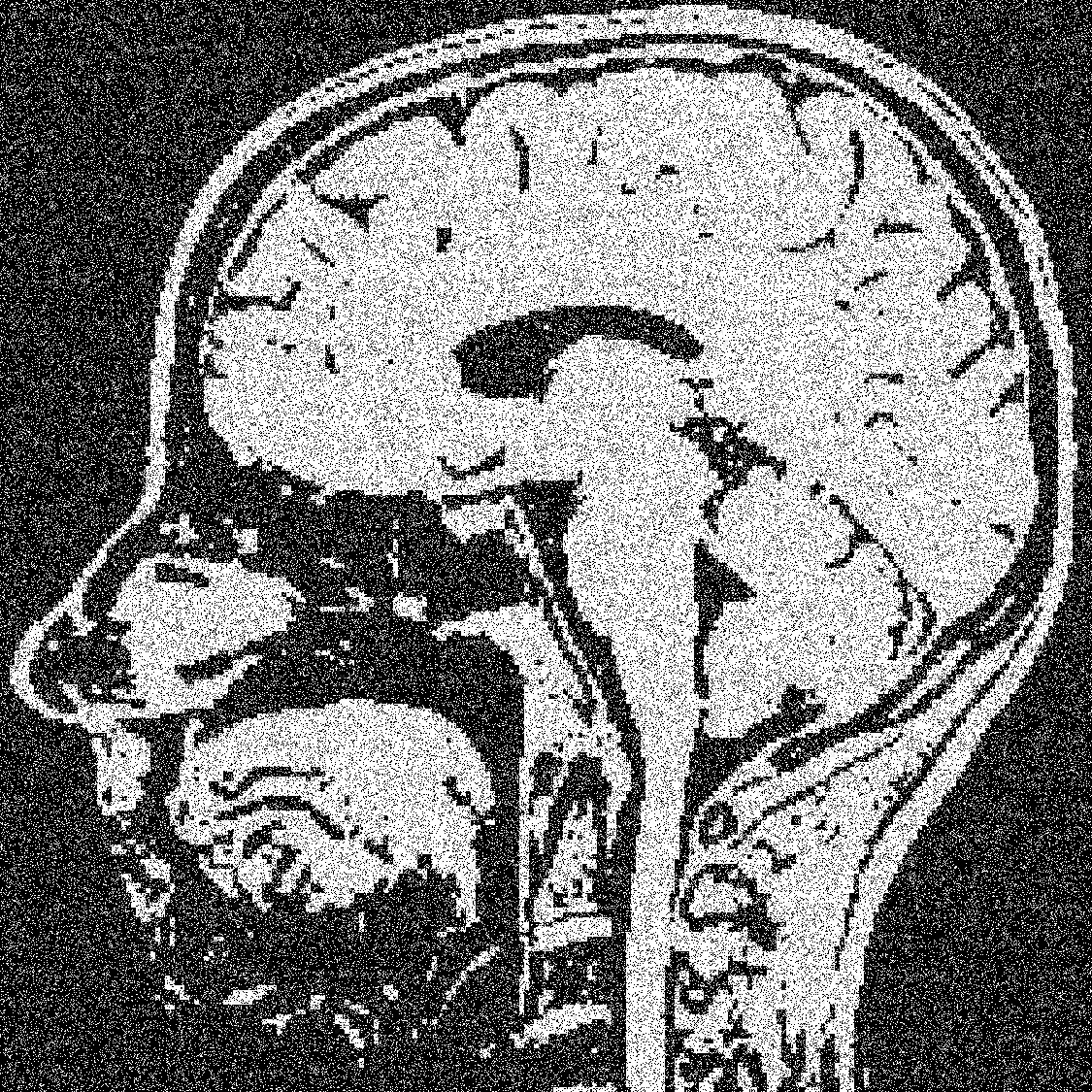}
\label{fig2}
\end{minipage}
\hspace{0.6cm}
\begin {minipage}{0.10\textwidth}
\includegraphics[width=\linewidth]{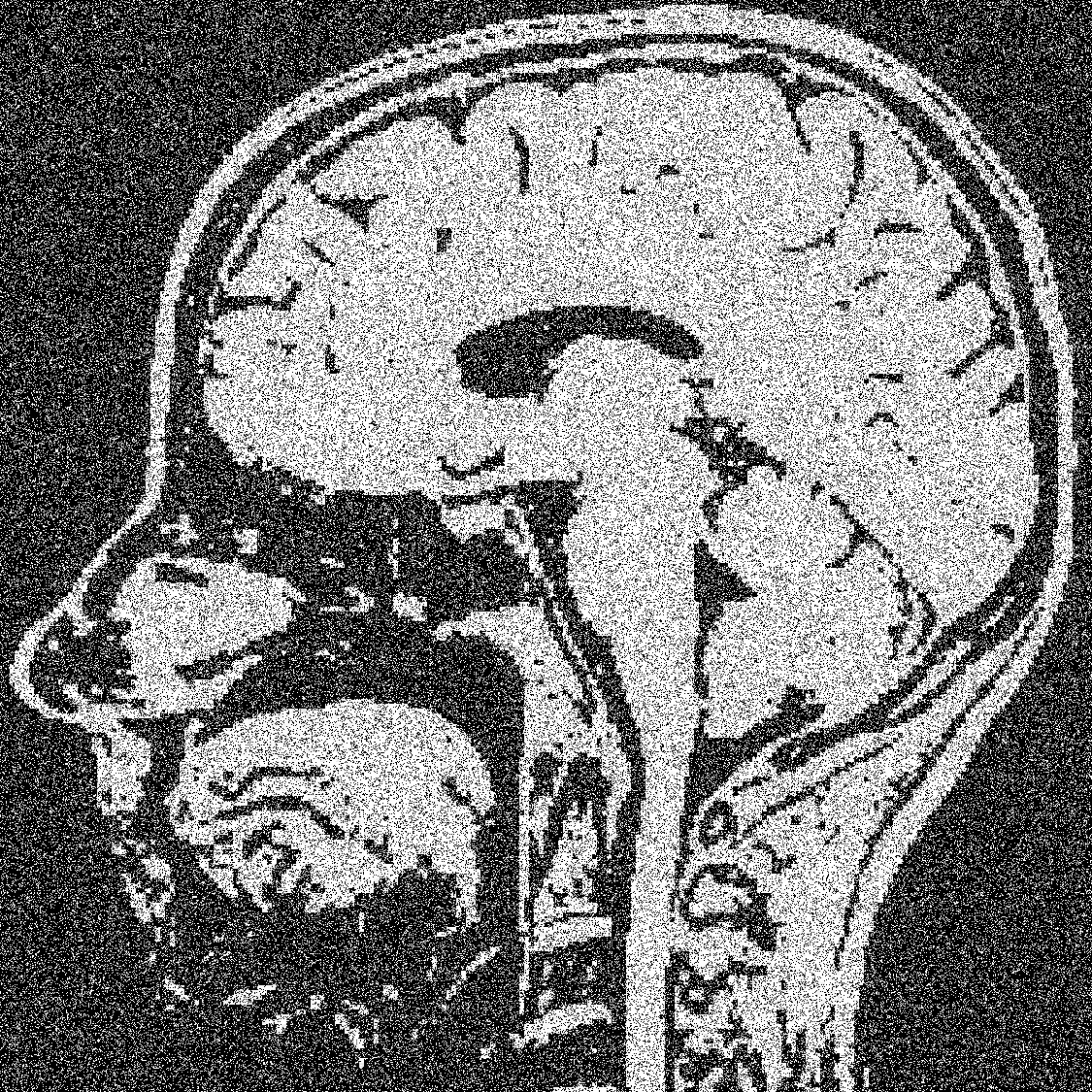}
\label{fig2}
\end{minipage}
\caption{MRI image with increasing level of noise}
\end{figure}

\begin{center}
	TABLE 2\\
	
	MSE FOR DIFFERENT FILTERS APPLIED TO FOUR NOISY VERSIONS OF MRI IMAGES  
\end{center}
\begin{center}
	\begin{tabular}{|c|c|c|c|}

		\hline
		
		Noise Ratio&Median Filter&Graph
		ASF&Hypergraph ASF\\
		
		\hline
		
		5 & 0.47 $(3\times3)$ & 0.77(2) & 0.23(2)\\
		10 & 0.84 $(3\times3)$ & 1.28(2) & 0.77(3)\\
		15 & 1.51 $(5\times5)$ & 2.24(2) & 1.89(3)\\
		20 & 1.84 $(5\times5)$ & 3.03(4) & 4.31(4)\\
		\hline
		
	\end{tabular}	
\end{center}

The implimentation is done in python using Open CV2 which run on intel core i3 processor with 4GB RAM.\\ 
\section{Conclusions }
Here we evaluate the effectiveness of the hypergraph based ASF filters. The results shows that hypergraph based ASF performs well with minimum error. It is observed that when increasing the scale of ASF, the mean square error is first decreasing, reaches a minimum value and then increasing. To verify quantitatively, we take four noisy versions of two binary images and apply ASF filters on them up to the scale 7. Comparisons of the performance of each filters in the different noisy versions of images are done. Mean square error in percentage is taken as the error measure. Error value in hypergraph based ASF increases for images with more than 10 \% of noise. But Better results can be obtained by applying ASF filter with different set of hyperedges which is left for future work. We are also developing algorithms for filters that can be applied effectively on gray scale images.






\begin{thebibliography}{1}
	
	\bibitem{1} Heijmans, Henk JAM and Christian Ronse, "The algebraic basis of mathematical morphology I. Dilations and erosions, \emph{Computer Vision, Graphics, and Image Processing}, vol. 50, no. 3, pp. 245-295, 1990.
	
	
	\bibitem{2}Gonzalez, Rafael C., and Richard E. Woods. Digital image processing, 2nd.\emph {SL: Prentice Hall}, 2002.
	
	\bibitem{3}Berge, Claude. "Hypergraphs: Combinations of Finite Sets." (1989).
	
	
	
	
	\bibitem{4}Henk Heijmans and L Vincent, Graph Morphology in Image Analysis, 1993.
	
	\bibitem{5}Vincent, Graphs and Mathematical Morphology, \emph{Signal Processing}, vol. 16, no. 4, pp. 365-388, 1989.
	
	
	
	\bibitem{6}Cousty, Jean, Laurent Najman, Fabio Dias, and Jean Serra, "Morphological filtering on graphs", \emph{Computer Vision and Image Understanding}, vol. 117, no. 4, pp. 370-385, 2013.
	\bibitem{7}Bloch, Isabelle, and Alain Bretto. "Mathematical morphology on hypergraphs: Preliminary definitions and results." \emph{Discrete Geometry for Computer Imagery,} Springer Berlin Heidelberg, 2011.
	\bibitem{8}Sebastian, V. Bino, A. Unnikrishnan, Kannan Balakrishnan, and P. B. Ramkumar, "Mathematical Morphology on Hypergraphs Using Vertex-Hyperedge Correspondence," \emph{ISRN Discrete Mathematics}, 2014. 
	\bibitem{9}Sebastian, V. Bino, A. Unnikrishnan, Kannan Balakrishnan, and P. B. Ramkumar, "Morphological filtering on hypergraphs," \emph{arXiv preprint arXiv:1402.4258}, 2014.
	\bibitem{10}Frank Y Shih, \emph{Image Processing and Mathematical Morphology: Fundamentals and Applications}, CRC press, 2010.
	\bibitem{11}Henk JAM Heijmans, Composing Morphological Filters, \emph{Image Processing, IEEE Transactions} vol. 6, no. 5, pp. 713-723, 1997.
	
	\bibitem{12}Jean Paul Serra, Image Analysis and Mathematical Morphology, 1982.
	
	\bibitem{13}Jean Cousty, Laurent Najman, and Jean Serra, "Some morphological operators in graph spaces",
	\emph{In Mathematical Morphology and Its Application to Signal and Image Processing}, pp 149–
	160, Springer, 2009.
	\bibitem{14}Isabelle Bloch and Alain Bretto. Mathematical morphology on hypergraphs, application
	to similarity and positive kernel. \emph{Computer vision and image understanding}, vol. 117, no. 4, pp. 342–354, 2013.
	
	
\end{thebibliography}
%

\end{document}